\documentclass[runningheads]{llncs}

\usepackage{amssymb}
\usepackage{subcaption, multirow, array}
\usepackage{booktabs}
\usepackage{xspace}
\usepackage{mathtools}
\usepackage{siunitx}
\usepackage{orcidlink}
\usepackage[misc]{ifsym}
\usepackage[normalem]{ulem}
\usepackage{censor}

\mathchardef\mhyphen="2D
\newcolumntype{C}{@{\extracolsep{6pt}}c@{\extracolsep{3pt}}}%
\newcolumntype{L}{@{\extracolsep{6pt}}l@{\extracolsep{3pt}}}%

\newcommand{\method}{StrokeTimer}
  
\makeatletter
\DeclareRobustCommand\onedot{\futurelet\@let@token\@onedot}
\def\@onedot{\ifx\@let@token.\else.\null\fi\xspace}

\newif\ifanonymize
\anonymizetrue    

\makeatother

\newcommand{\subpara}[1]{\vspace{3pt} \noindent \textbf{#1}}
\usepackage[T1]{fontenc}
\usepackage{graphicx,verbatim}
\usepackage{hyperref}
\usepackage{color}

\begin{document}
%
\title{StrokeTimer: Robust Representation Learning for Ischemic Stroke Onset-Time Estimation from Non-contrast CT}
\titlerunning{StrokeTimer: Stroke Onset Time Estimation from NCCT}
%

\author{
Weiru Wang\inst{1,2} \and
Susanne G.H. Olthuis\inst{3} \and
Elizaveta Lavrova\inst{4} \and
Robert J. van Oostenbrugge\inst{3} \and
Charles B.L.M. Majoie\inst{5} \and
Wim H. van Zwam\inst{6} \and
Ruisheng Su\inst{1} 
}

\authorrunning{Wang et al.}

\institute{
Department of Biomedical Engineering, Eindhoven University of Technology, Eindhoven, The Netherlands
\and
Graduate School of Life Sciences, Utrecht University, Utrecht, The Netherlands
\and
Department of Neurology, Maastricht University Medical Centre+, Maastricht, The Netherlands
\and
Precision Medicine Department, GROW Research Institute for Oncology and Reproduction, Maastricht University,  Maastricht, The Netherlands
\and
Department of Radiology and Nuclear Medicine, Amsterdam University Medical Centre, Amsterdam, The Netherlands
\and
Department of Radiology and Nuclear Medicine, Maastricht University Medical Centre+, Maastricht, The Netherlands
}

\maketitle              
\begin{abstract}
Ischemic stroke is a major global disease. Treatment decisions are highly time-sensitive, as eligibility for reperfusion therapies relies on the interval between stroke onset and intervention. However, the true onset time is often uncertain in clinical practice, necessitating imaging-based assessment of tissue age as a surrogate marker. Early ischemic changes on routinely acquired non-contrast CT (NCCT) are often subtle, and real-world clinical datasets exhibit pronounced onset-time class imbalance and center-scanner-related heterogeneity. In this work, we propose \method{}, a fully automated framework for onset-time estimation in acute ischemic stroke. \method{} integrates self-supervised disentanglement learning with energy-guided contrastive learning to capture subtle ischemic patterns while addressing long-tailed data distributions under acquisition variability. Onset time is categorized into three clinically relevant windows (<4.5\,h, 4.5--6\,h, and >6\,h). Experimental results on a large multi-center NCCT dataset from two national cohorts, MR CLEAN Registry and MR CLEAN LATE, show that \method{} achieves a macro AUC of 0.69 and a macro F1-score of 0.57, improving the strongest baseline by nearly 50\% ($p < 0.005$). In this realistic, challenging setting, representative baseline approaches exhibit near-chance macro performance. Model explanations further highlight subtle gray–white matter blurring and hypodense regions consistent with established radiological biomarkers.
These findings demonstrate the potential of \method{} to support treatment decision-making in acute ischemic stroke. Code is available at \url{https://github.com/BrainVas/StrokeTimer}.

\end{abstract}
\section{Introduction}

Stroke is the second leading cause of death and long-term disability worldwide, with over 3.3 million deaths annually~\cite{Feigin2021GBDStroke,WSO}. In acute ischemic stroke, therapeutic strategies such as intravenous thrombolysis and mechanical thrombectomy are governed by time-dependent eligibility criteria. Therefore, determining whether a patient falls within a treatable time window is key to clinical decision-making. However, stroke onset is frequently unwitnessed, creating uncertainty in the true time of onset and thus the need for imaging-based indicators of tissue viability. 

While MRI provides high sensitivity for early infarction, routinely available non-contrast CT (NCCT) remains the first-line imaging modality in emergency workflows due to its broad availability and rapid acquisition~\cite{MRI_stroke2,provenzale2003assessment,MRI_stroke1}. Estimating onset time from NCCT, however, is challenging because early ischemic alterations are often subtle to distinguish from normal variation~\cite{ncct_subtle1,provenzale2003assessment,Sun2024MultiGrained}. Apart from this intrinsic difficulty, real-world imaging datasets exhibit substantial heterogeneity arising from differences in scanner vendors, acquisition protocols, and reconstruction settings. Furthermore, clinical registries constrained by time-based treatment criteria often come with high class imbalance across time windows, forming long-tailed data distributions~\cite{Albers2018DEFUSE3,MRCLEAN_Registry}.

To address imaging heterogeneity, domain adaptation and disentanglement learning have been explored for scanner harmonization~\cite{Zuo2021Harmonization}. Domain adaptation methods typically reduce style variability through feature alignment or appearance normalization. However, in the presence of high diversity of imaging styles, conventional strategies, such as domain adversarial networks~\cite{Kanakasabapathy2021,Zhao2018AMDA} or CycleGAN~\cite{Yan2019UnetGAN}, often struggle to achieve stable cross-domain alignment. Moreover, generative models may inadvertently introduce spurious features or suppress subtle pathological cues, which is undesirable in safety-critical clinical tasks~\cite{Wang2025GenAI}. 

Long-tailed learning is commonly addressed with loss engineering strategies (e.g., Focal loss~\cite{Lin2020FocalLoss}) or decoupled training schemes that separate representation learning from classifier rebalancing (e.g., MARC~\cite{wang2023margin}). In parallel, self-supervised learning (SSL) has shown promise for learning robust, transferable representations that partially alleviate long-tailed effects. However, current approaches typically address either data imbalance or domain variability in isolation.

To capture onset time-related subtle changes and estimate onset time from medical imaging, existing computational approaches rely on manual voxel-level lesion mask delineation for classification or regression, which can be labor-intensive and particularly unreliable in the hyperacute stage~\cite{regression,Sun2024MultiGrained,vorberg2025leveraging,van2025machine}. 
In this work, we introduce \method{}, the first fully automatic framework for onset time classification in acute ischemic stroke from routine non-contrast CT (NCCT) that simultaneously accounts for data imbalance and domain variation without requiring prior manual delineation of stroke lesions.

\section{Methods}
\label{sec:methods_overview}
Given a skull-stripped NCCT volume \( x \in \mathbb{R}^{1 \times D \times H \times W} \) from a patient with acute ischemic stroke, the goal is to estimate the true stroke onset time window label \( y \in \{0~(<4.5\text{h}), 1~(4.5-6\text{h}), 2~(>6\text{h})\} \).

\begin{figure}[!t]
    \centering
    \includegraphics[width=\textwidth]{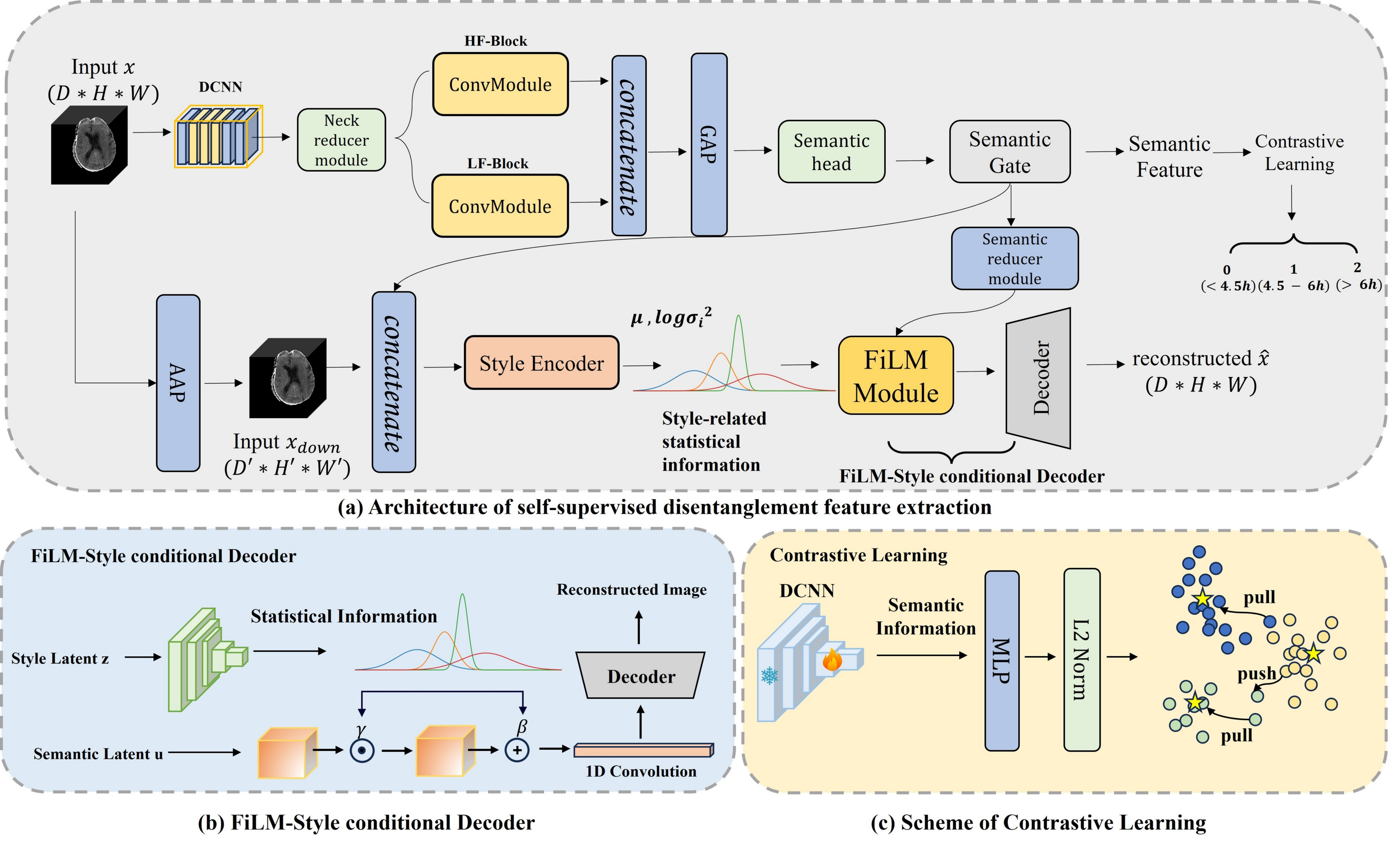}
    \caption{Overview of StrokeTimer: a reconstruction-based self-supervised disentanglement framework with independent semantic and style branches, followed by contrastive refinement.}
    \label{fig:method}
\end{figure}

We propose \method{}, a two-stage framework (Fig.~\ref{fig:method}). The first stage involves a self-supervised semantic-style disentanglement module that learns a style-invariant semantic representation by reconstructing the input using a FiLM-conditioned decoder~\cite{Perez2018FiLM}. The second stage refines the embeddings solely from semantic branch using an Energy-guided Contrastive Mean-Shift (ECMS) module, which combines prototype-based~\cite{Li2021PCL,Snell2017ProtoNet} contrastive learning with frequency-aware prototype refinement to handle class imbalance and subtle inter-class differences.

\subsection{Semantic-Style Disentanglement}
\label{sec:methods_disentangle}
We use a 3D ResNeXt backbone to extract volumetric features from NCCT scans, with two convolutional branches that capture both high- and low-frequency components with different kernel sizes. The resulting features are concatenated and globally pooled into compact semantic features \( s \in \mathbb{R}^{d_s} \). A channel-wise semantic gate modulates spatial feature maps conditioned on \( s \), yielding a compact semantic content map \( U \).

To model acquisition-dependent style variations, we introduce a VAE-style~\cite{Kingma2019VAE} encoder that combines downsampled input by adaptive average pooling with semantic features, mapped to a Gaussian posterior with parameters \( (\mu, \log \sigma^2) \). A style latent variable \( z \) is sampled via reparameterization.


The FiLM-conditioned decoder then reconstructs the input by modulating semantic features using the style latent variable \( z \). The modulation is achieved using learned scaling and bias parameters \( \gamma(z) \) and \( \beta(z) \):

\begin{equation}
\tilde{U} = (1 + \gamma(z)) \odot U + \beta(z)
\end{equation}

The disentanglement stage is optimized with a joint orthogonality loss~\cite{Navab2020Fairness} to realize semantic-style independence. 
\begin{equation}
\mathcal{L}_{\mathrm{dis}} = \lambda_{\mathrm{rec}}\|\hat{x} - x\|_1 + \lambda_{\mathrm{KL}} \mathrm{KL}(q(z|x) \| p(z)) + \lambda_{\mathrm{cyc}}\|\tilde{z} - z\|_2^2 + \lambda_{\mathrm{orth}}\|S^\top Z\|_F^2
\end{equation}
where $\lambda_{\mathrm{rec}}$, $\lambda_{\mathrm{KL}}$, $\lambda_{\mathrm{cyc}}$, and $\lambda_{\mathrm{orth}}$ are scalar weights controlling the contribution of each loss term, enforcing faithful reconstruction, regularizing the latent distribution, preserving cycle-consistent latent representations, and explicitly promoting semantic--style disentanglement by enforcing orthogonality between semantic and imaging factors.

\subsection{Energy-Guided Contrastive Mean-Shift (ECMS)}
\label{sec:methods_ecms}
We refine semantic embeddings using ECMS, which integrates prototype-based contrastive learning with frequency-aware prototype refinement. Semantic embeddings are projected and $\ell_2$-normalized to obtain unified feature vectors $v_i$. For each class, a prototype $c_k$ is maintained and optimized via prototype contrastive across-entropy loss~\cite{Li2021PCL}:

\begin{equation}
\mathcal{L}_{\mathrm{proto}} = -\frac{1}{N} \sum_{i=1}^N \log \frac{\exp(v_i^\top c_{y_i}/\tau)}{\sum_{k=1}^K \exp(v_i^\top c_k/\tau)}
\end{equation}

To enhance class discrimination, we update prototypes using a similarity-weighted mean-shift strategy:
\begin{equation}
m_k = \sum_{i:y_i=k} w_i\, v_i,
\end{equation}
where \(v_i\) denotes the $\ell_2$-normalized embedding of sample \(i\), and the weights
\begin{equation}
w_i = \frac{\exp\!\left(\langle v_i, c_k \rangle / \tau_{\mathrm{ms}}\right)}
{\sum_{j:y_j=k} \exp\!\left(\langle v_j, c_k \rangle / \tau_{\mathrm{ms}}\right)}
\end{equation}
are computed based on cosine similarity with the current class prototype \(c_k\),
where \(\tau_{\mathrm{ms}}\) controls the sharpness of the kernel, assigning higher
weights to samples more aligned with the prototype and yielding a stable estimate
of the local cluster center. The prototype is updated via:
\begin{equation}
\mathbf{c}_k \leftarrow (1 - \alpha_k)\mathbf{c}_k + \alpha_k\mathbf{m}_k,
\quad \alpha_k = (1 - m_k)\gamma_k,
\end{equation}
where $\gamma_k \propto (1/f_k)^{\delta}$ scales the update rate inversely with class frequency with $\delta > 0$ a hyperparameter, enabling faster adaptation for tail classes, and $\mathbf{m}_k$ represents the batch-level class center with $m_k = \|\mathbf{m}_k\|$ its scalar magnitude. 

To further account for long-tailed data distribution, we introduce a frequency-based prior $w_k \propto (1/f_k)^{\gamma}$, derived from global class frequencies $f_k$ with hyperparameter $\gamma > 0$. The final ECMS objective is defined as
\begin{equation}
\mathcal{L}_{\mathrm{ECMS}} =
\frac{1}{N}\sum_{i=1}^N w_{y_i}\,\mathcal{L}_{\mathrm{proto}}^{(i)},
\label{eq:lecms}
\end{equation}
where \( y_i \) denotes the class label of sample \( i \),
\( w_{y_i} \) is the corresponding class-level weight, and
\(\mathcal{L}_{\mathrm{proto}}^{(i)}\) represents the prototype contrastive
loss of sample \(i\).

\begin{table}[t]
\centering
\caption{Performance comparison of \method{} against representative long-tailed classification baselines. Statistical significance versus random guessing is assessed using a binomial test (\textbf{\emph{p-value}}).}
\label{tab:loss_comparison}
\begin{tabular}{l|l|cccccc}
\hline
\textbf{Type} & \textbf{Method} 
& \textbf{ACC} & \textbf{PRE} & \textbf{REC} & \textbf{F1} & \textbf{AUC} & \textbf{\emph{p-value}} \\
\hline
Decoupling 
& MARC ~\cite{wang2023margin}       
& 0.34 & 0.23 & 0.33 & 0.27 & 0.38 & 5.17e-01 \\
& $\tau$-Norm ~\cite{Kang2020Decoupling}
& 0.34 & 0.11 & 0.33 & 0.17 & 0.51 & 5.17e-01 \\
\hline
Loss       
& Focal Loss~\cite{Lin2020FocalLoss}  
& 0.34 & 0.11 & 0.33 & 0.17 & 0.44 & 5.17e-01 \\
& BSCE ~\cite{Ren2020BalancedSoftmax}       
& 0.48 & 0.32 & 0.49 & 0.39 & 0.65 & 6.81e-02 \\
\hline
CL         
& MulSupCon ~\cite{Zhang2024MLSupCon}  
& 0.37 & 0.29 & 0.37 & 0.24 & \textbf{0.73} & 4.11e-01 \\
& SoftCon ~\cite{Wang2024MLGSoft}    
& 0.37 & 0.37 & 0.37 & 0.29 & 0.68 & 4.11e-01 \\
\hline
Ours         
& StrokeTimer w/o ECMS 
              & 0.52 & 0.53 & 0.51 & 0.52 & 0.65 & 3.11e-02 \\
& StrokeTimer 
              & \textbf{0.59} & \textbf{0.58} & \textbf{0.59} & \textbf{0.57} & 0.69 & \textbf{4.52e-03} \\
\hline
\end{tabular}
\end{table}
\begin{table*}[t]
\centering
\caption{Per-class performance comparison of different methods.}
\label{tab:perclass_all_methods}
\small
\resizebox{\textwidth}{!}{%
\begin{tabular}{lcccc|cccc|cccc}
\toprule
\textbf{Method}
& \multicolumn{12}{c}{\textbf{Stroke Onset Time Windows}} \\
\cmidrule(lr){2-13}
& \multicolumn{4}{c}{$< 4.5$ h}
& \multicolumn{4}{c}{4.5--6 h}
& \multicolumn{4}{c}{$> 6$ h} \\
\cmidrule(lr){2-5} \cmidrule(lr){6-9} \cmidrule(lr){10-13}
& PRE & REC & F1 & AUC
& PRE & REC & F1 & AUC
& PRE & REC & F1 & AUC \\
\midrule
MARC~\cite{wang2023margin}
& 0.27 & 0.40 & 0.32 & 0.34
& 0.43 & \textbf{0.60} & 0.50 & 0.49
& 0.00 & 0.00 & 0.00 & 0.31 \\

BSCE~\cite{Ren2020BalancedSoftmax}
& 0.47 & 0.70 & 0.56 & 0.71
& 0.00 & 0.00 & 0.00 & 0.53
& 0.50 & \textbf{0.78} & 0.61 & 0.71 \\

$\tau$-Norm ~\cite{Kang2020Decoupling}
& 0.34 & 1.00 & 0.51 & 0.45
& 0.00 & 0.00 & 0.00 & 0.53
& 0.00 & 0.00 & 0.00 & 0.56 \\

Focal Loss ~\cite{Lin2020FocalLoss}
& 0.34 & \textbf{1.00} & 0.51 & 0.44
& 0.00 & 0.00 & 0.00 & 0.60
& 0.00 & 0.00 & 0.00 & 0.29 \\

MulSupCon~\cite{Zhang2024MLSupCon}
& 0.36 & \textbf{1.00} & 0.53 & 0.77
& 0.00 & 0.00 & 0.00 & \textbf{0.70}
& 0.50 & 0.11 & 0.18 & 0.73 \\

SoftCon~\cite{Wang2024MLGSoft}
& 0.40 & \textbf{0.89} & 0.55 & \textbf{0.78}
& 0.20 & 0.11 & 0.14 & 0.57
& 0.50 & 0.11 & 0.18 & 0.68 \\ \hline

StrokeTimer w/o ECMS (Ours)
& \textbf{0.67} & 0.60 & 0.63 & 0.67
& \textbf{0.56} & 0.50 & \textbf{0.53} & 0.65
& 0.36 & 0.44 & 0.40 & 0.62 \\

StrokeTimer (Ours)
& 0.57 & 0.80 & \textbf{0.67} & 0.75
& 0.50 & 0.30 & 0.38 & 0.54
& \textbf{0.67} & 0.67 & \textbf{0.67} & \textbf{0.78} \\
\bottomrule
\end{tabular}
}
\end{table*}

\section{Experiments and Results}
\subsection{Experimental Setup}
\subpara{Datasets}
A total of 1,686 non-contrast CT (NCCT) brain scans were retrospectively collected from the MR CLEAN Registry and MR CLEAN-LATE datasets, covering eighteen medical centers. The cohort consisted of 1,531 cases with onset $<4.5$\,h, 72 cases between 4.5--6\,h, and 83 cases $>6$\,h. The dataset was randomly split into 1,626 training scans, 30 validation scans, and 30 testing scans. Stratified sampling was applied to ensure a uniform class distribution across the validation and testing sets. All volumes were intensity-normalized to $[-1,1]$, padded and resampled to $48 \times 256 \times 256$, and skull-stripped using CT-BET~\cite{akkus2020brain}.

\subpara{Implementation Details.}
All models were implemented in PyTorch. The backbone network was a ResNeXt3D encoder equipped with a disentanglement module and an ECMS head, followed by a cosine classifier. Optimization was performed using SGD with momentum 0.9 and weight decay $3 \times 10^{-4}$. Training proceeded in two stages. The initial stage was trained with a base learning rate of $3\times10^{-4}$, while the second stage used a learning rate reduced to 0.02 times the initial value. A 10-epoch warm-up strategy was applied prior to a step-wise decay schedule. The disentanglement stage was trained for 80 epochs, followed by ECMS training from epoch 81 to 150. The batch size was set to 8. For the ECMS module, we set the energy-based weight exponent $\gamma=0.1$, mean-shift temperature $\tau_{ms}=0.2$, and frequency scaling exponent $\delta=1.0$. All experiments were conducted on a single NVIDIA H200 GPU, and one full two-stage training required approximately 9 hours.

\subpara{Baselines.}
\method{} was compared with six long-tailed classification methods (Focal Loss~\cite{Lin2020FocalLoss}, Balanced Softmax~\cite{Ren2020BalancedSoftmax}, $\tau$-Norm~\cite{Kang2020Decoupling}, MulSupCon~\cite{Zhang2024MLSupCon}, SoftCon~\cite{Wang2024MLGSoft} and MARC~\cite{wang2023margin}) and four SSL approaches (MoCo~\cite{He2020MoCo}, SimCLR~\cite{Chen2020SimCLR}, VoCo~\cite{Wu2024VoCo}, and MAE~\cite{He2022MAE}).

\subpara{Evaluation Metrics.}
We conducted qualitative and quantitative evaluations on the test set. Quantitative metrics included accuracy, macro-averaged precision, recall, F1-score, and AUC. Macro metrics were obtained by averaging per-class results.

\begin{figure*}[!t]
    \centering
    \includegraphics[width=\textwidth]{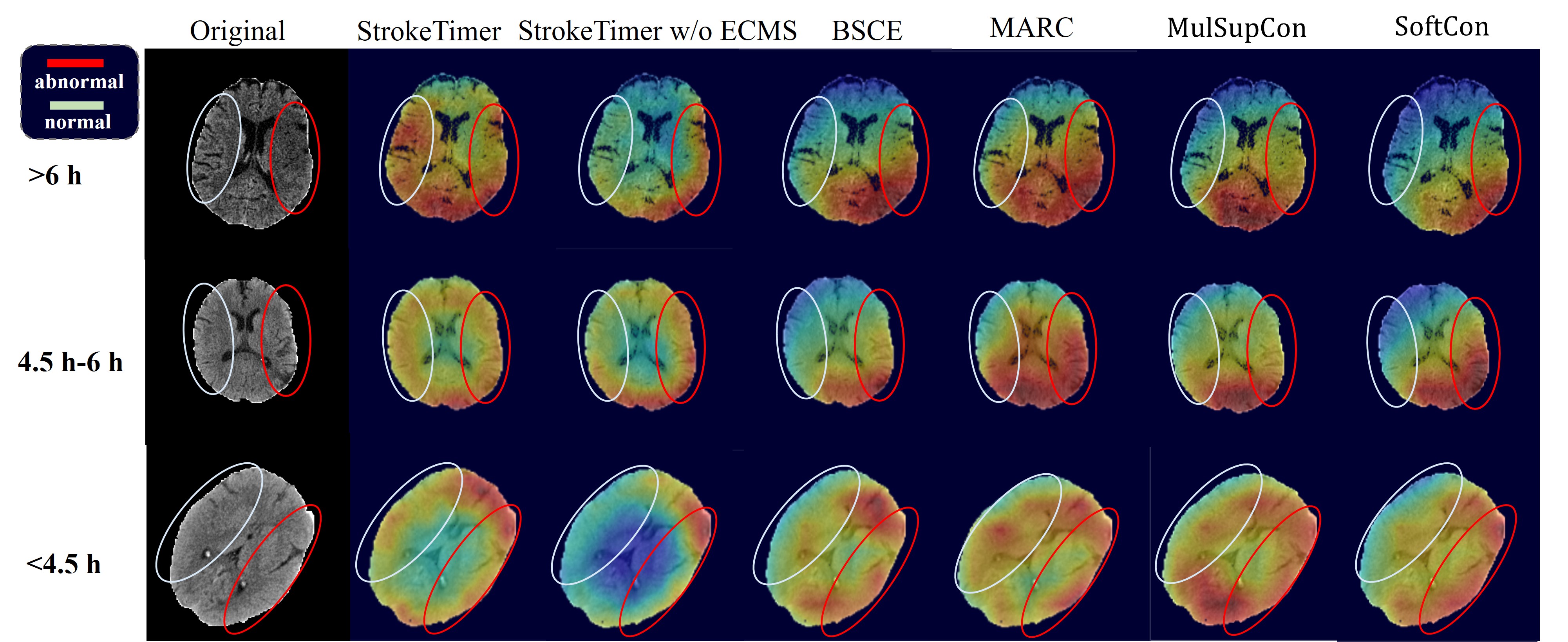}
    \caption{Grad-CAM++ heatmap comparison between \method{} and top 4 representative baseline approaches across onset time categories.}
    \label{fig:ablation}
\end{figure*}

\begin{table}[!t]
\centering
\caption{Performance comparison between representative SSL methods and the proposed \method{}. Statistical significance versus random guessing is assessed using a binomial test (\textbf{\emph{p-value}}).}
\label{tab:self-supervised}
\begin{tabular}{l|cccccc}
\hline
\textbf{Method} 
& \textbf{ACC} 
& \textbf{PRE} 
& \textbf{REC} 
& \textbf{F1} 
& \textbf{AUC} 
& \textbf{\emph{p-value}} \\
\hline
SimCLR~\cite{Chen2020SimCLR} 
& 0.31 & 0.21 & 0.30 & 0.23 & 0.47 & 6.70e-01 \\
MoCo ~\cite{He2020MoCo}
& 0.28 & 0.26 & 0.27 & 0.19 & 0.45 & 8.01e-01 \\
VoCo ~\cite{Wu2024VoCo}
& 0.28 & 0.27 & 0.28 & 0.27 & 0.44 & 8.01e-01 \\
MAE ~\cite{He2022MAE}
& 0.31 & 0.33 & 0.16 & 0.10 & 0.33 & 6.70e-01 \\
\hline
\method{} w/o ECMS 
& \textbf{0.52} 
& \textbf{0.53} 
& \textbf{0.51} 
& \textbf{0.52} 
& \textbf{0.65} 
& \textbf{3.11e-02} \\

\hline
\end{tabular}
\end{table}

\subsection{Results}
\subpara{Quantitative Analysis.}
To evaluate the effectiveness of \method{} in addressing severe class imbalance in onset-time distributions, we compare it with common long-tailed learning strategies across the three time windows. Tables~\ref{tab:loss_comparison} and~\ref{tab:perclass_all_methods} summarize both aggregate and per-class performance.

At the aggregate level, \method{} achieves the best macro performance, with an accuracy of 0.59 and a macro F1-score of 0.57, significantly outperforming all baselines ($p < 0.005$). Compared to the strongest baseline method, \method{} improves the macro F1-score by nearly twofold, indicating improved classification consistency under high class imbalance.

At the per-class level, \method{} demonstrates balanced discriminative capability across clinically relevant time windows. It achieves AUCs of 0.75 and 0.78 in the $<$4.5\,h and $>$6\,h groups, respectively, outperforming competing long-tailed approaches. Importantly, while most baselines show near-zero precision and recall in the challenging 4.5--6\,h intermediate window, \method{} maintains stable performance (F1 = 0.53, PRE = 0.56), suggesting improved sensitivity and robustness in detecting subtle temporal transitions.

To assess whether the performance improvements arise from the proposed domain-specific design or from the intrinsic generalization capability of SSL, we further compare \method{} with representative SSL frameworks. As shown in Table~\ref{tab:self-supervised}, SSL approaches show near-chance performance (ACC = $[0.28, 0.31]$) when applied to multi-center NCCT stroke data, despite their strong results on natural image benchmarks. In contrast, \method{} w/o ECMS achieves a statistically significantly higher accuracy (ACC = 0.52 vs. $[0.28, 0.31]$, $p < 0.05$), underscoring the added value of domain-specific architectural design when addressing long-tailed distributions and acquisition variability in medical imaging.

\begin{figure}[!t]
    \centering
    \includegraphics[width=\linewidth]{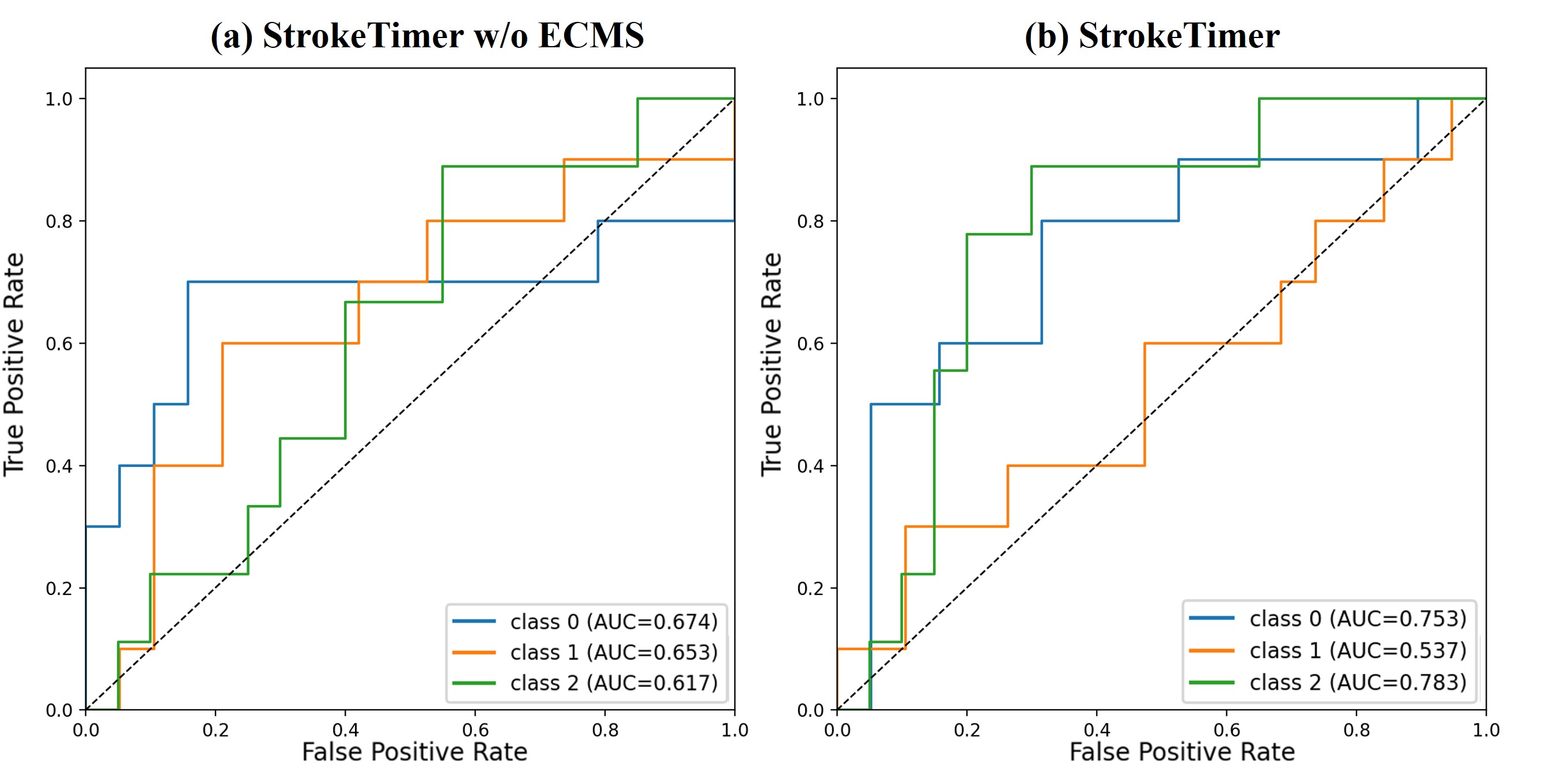}
    \caption{Comparison between the disentanglement-only and full \method{} (Disentanglement+ECMS) in terms of ROC curves }
    \label{fig:cm_roc_comparison}
\end{figure}

\subpara{Qualitative Analysis.}
To qualitatively evaluate \method{}, we visualized Grad-CAM++~\cite{gradcam} attention maps and compared them with those of representative long-tailed learning approaches (Fig.~\ref{fig:ablation}). While baseline methods tend to produce diffuse or anatomically irrelevant activations, \method{} generates more coherent and anatomically plausible attention patterns. With ECMS, attention maps become more focused and symmetric, highlighting cortical hypoattenuation, early loss of gray-white matter differentiation, and contralateral hemispheric asymmetry, which correspond to clinical biomarkers on NCCT. To further assess the contribution of ECMS, we compared ROC curves across training stages (Fig.~\ref{fig:cm_roc_comparison}). Incorporating ECMS improves discrimination performance for the two boundary classes, whereas performance for the transitional 4.5-6h class shows a slight reduction.

\section{Discussion and Conclusion}
\method{} demonstrates robust performance in classifying stroke onset time windows from NCCT without the need for manual lesion delineation, effectively addressing long-tailed distributions and style variability. It outperforms common baselines in both long-tailed classification and self-supervised learning for NCCT-based onset time classification. Ablation studies reveal that \method{} distinguishes subtle progressive changes and highlights clinically corresponding features, indicating that explicit disentanglement learning serves as the key performance driver by separating semantic and style features. Integration with self-supervised learning yields more generalizable representations, mitigating head-class dominance in long-tailed distributions.

The ECMS module introduces a clinically interpretable performance trade-off. While the AUC of the intermediate 4.5–6\,h class decreases modestly (0.653 to 0.537), ECMS substantially improves discrimination of the two clinically critical classes (<4.5 h: 0.674 to 0.753; >6 h: 0.617 to 0.783). From a clinical perspective, accurate identification of these boundary windows is particularly critical, as they directly inform decisions regarding thrombolysis eligibility or deferral of reperfusion therapy. In contrast, patients within the transitional window (4.5-6\,h) typically undergo additional perfusion imaging per current guidelines. An NCCT-based estimate may therefore help triage the need for perfusion imaging. However, the limited number of training samples in the 4.5-6\,h group (n=72) constrains achievable generalization. The prototype-based contrastive mechanism promotes tighter class separation, which may increase ambiguity for samples exhibiting mixed ischemic features. This behavior reflects the inherent diagnostic uncertainty of this transitional window faced by clinicians.

In conclusion, we presented \method{}, establishing the first fully automatic benchmark for NCCT-based onset estimation in acute ischemic stroke. \method{} addresses both long-tailed class imbalance and image heterogeneity without requiring any manual input. Experimental results on a large multi-center clinical dataset demonstrate robust performance, revealing its potential to support treatment decision-making in acute ischemic stroke.

%
%
\subsubsection{Acknowledgement} 
We thank the MR CLEAN Registry and MR CLEAN Late investigators for their contribution.
\subsubsection{Disclosure of Interests}  The authors have no competing interests to declare that are relevant to the content of this article.

\bibliographystyle{splncs04}
\bibliography{references}

\end{document}